\documentclass{article}

\usepackage[final]{airov25}

\usepackage[utf8]{inputenc} % allow utf-8 input
\usepackage[T1]{fontenc}    % use 8-bit T1 fonts
\usepackage{booktabs}       % professional-quality tables
\usepackage{amsfonts}       % blackboard math symbols
\usepackage{nicefrac}       % compact symbols for 1/2, etc.
\usepackage{microtype}      % microtypography
\usepackage{xcolor}         % colors
\usepackage{graphicx}       % graphicx
\usepackage{lipsum}

\title{Towards a Framework for Supporting the Ethical and Regulatory Certification of AI Systems}

\author{
  \textbf{Fabian Kovac$^{1}$ \quad Sebastian Neumaier$^{1}$ \quad Timea Pahi$^{1}$ \quad Torsten Priebe$^{1}$} \\
  \textbf{Rafael Rodrigues$^{2}$ \quad Dimitrios Christodoulou$^{2}$ \quad Maxime Cordy$^{3}$ \quad Sylvain Kubler$^{3}$} \\
  \textbf{Ali Kordia$^{3}$ \quad Georgios Pitsiladis$^{4}$ \quad John Soldatos$^{4}$ \quad Petros Zervoudakis$^{4}$} \\
  $^{1}$St. Pölten University of Applied Sciences \quad $^{2}$Digital for Planet - D4P\\ $^{3}$University of Luxembourg \quad $^{4}$Netcompany-Intrasoft S.A.\\
  \texttt{\{fabian.kovac, sebastian.neumaier, timea.pahi, torsten.priebe\}@fhstp.ac.at}\\
  \texttt{\{rafael.rodrigues, dimitrios.christodoulou\}@digital4planet.org}\\
  \texttt{\{maxime.cordy, sylvain.kubler, ali.kordia\}@uni.lu}\\
  \texttt{\{georgios.pitsiladis, john.soldatos, petros.zervoudakis\}@netcompany.com}
}

\begin{document}

\maketitle

\begin{abstract}
    Artificial Intelligence has rapidly become a cornerstone technology, significantly influencing Europe's societal and economic landscapes. However, the proliferation of AI also raises critical ethical, legal, and regulatory challenges.
    The CERTAIN (Certification for Ethical and Regulatory Transparency in Artificial Intelligence) project addresses these issues by developing a comprehensive framework that integrates regulatory compliance, ethical standards, and transparency into AI systems. 
    %Through semantic Machine Learning Operations (MLOps), ontology-driven data lineage tracking, and regulatory operations (RegOps) workflows, CERTAIN establishes foundational mechanisms ensuring AI systems' accountability, fairness, and sustainability.
    In this position paper, we outline the methodological steps for building the core components of this framework. Specifically, we present:
    (i) semantic Machine Learning Operations (MLOps) for structured AI lifecycle management,
    (ii) ontology-driven data lineage tracking to ensure traceability and accountability, and
    (iii) regulatory operations (RegOps) workflows to operationalize compliance requirements.
    By implementing and validating its solutions across diverse pilots, CERTAIN aims to advance regulatory compliance and to promote responsible AI innovation aligned with European standards.
\end{abstract}

\section{Introduction and Motivation}
\label{sec:introduction_and_motivation}
Trustworthy and regulatory-compliant artificial intelligence (AI) is paramount in shaping Europe's digital future. As AI technologies proliferate across various sectors, there emerges a critical need to ensure these systems are not only innovative but also transparent, ethical, and aligned with evolving EU regulatory frameworks. Recent work tracing the evolution of EU data regulations highlights the increasing complexity and breadth of regulatory reach \cite{pathak2024}, extending beyond personal data to include non-personal and business data, and encompassing new actors such as AI system developers, data intermediaries, and manufacturers of connected products. Within this context, the CERTAIN (Certification for Ethical and Regulatory Transparency in Artificial Intelligence) project\footnote{\href{https://certain-project.eu/}{https://certain-project.eu/}} positions itself as a pioneering initiative addressing these critical gaps. It seeks to build comprehensive mechanisms for regulatory transparency and certification. By creating tools, processes, and guidelines for stakeholders to achieve regulatory compliance and sustainability objectives, CERTAIN fosters trust, transparency, and responsible innovation.

This positional paper introduces CERTAIN, articulating its objectives, approach, and expected impacts. Furthermore, the paper presents the project's conceptual architecture, highlighting foundational components designed to streamline compliance validation processes and certification activities.

\section{Project Overview}
The CERTAIN project aims to create a cohesive and compliant ecosystem for AI stakeholders and includes 19 diverse partners representing academia and industry, spanning multiple European countries. The project explicitly targets adoptability and real-world applicability by validating its outcomes through seven distinct pilots across critical sectors: biometrics, health, energy, human resources, finance, and IT.

CERTAIN's main objectives include:
\begin{itemize}
    \item Ensuring traceability and transparency of AI systems using advanced semantic technologies.
    \item Producing multidisciplinary legal, ethical, and social guidelines to support compliance.
    \item Designing and demonstrating tools for data space providers to ensure regulatory compliance and minimize energy consumption.
    \item Developing realistic test methods and synthetic data generation techniques to assess and improve AI system compliance with regulatory requirements.
    \item Creating templates for AI certification processes tailored to various application domains.
\end{itemize}

Two critical components underpin CERTAIN's technical and regulatory compliance infrastructure:

\textbf{Semantic MLOps and Ontology Development} (cf. Section \ref{sec:semanticframework}) 
%provides the semantic logic and content necessary for compliance validation. It encompasses 
focuses on the development of ontologies for semantic data integration, automatic testing methods, and detailed certification scheme templates. 
By leveraging semantic web technologies and Knowledge Graphs, this component ensures rigorous and comprehensive validation against regulatory standards.

\textbf{Infrastructure and Compliance Mechanisms} (cf. Section \ref{sec:infracompliance}) focuses on building the foundational infrastructure, connectors, and compliance tools necessary for seamless regulatory adherence. This includes developing a data lineage connector for transparency and designing robust compliance checks against various EU regulatory frameworks.

\section{CERTAIN Framework}
The framework is designed to not only support technical implementation but also serve as a narrative tool, illustrating how CERTAIN enables stakeholders across the AI value chain to efficiently navigate complex regulatory landscapes and achieve certification-ready standards.

\begin{figure}[h!]
    \centering
    \includegraphics[width=\linewidth]{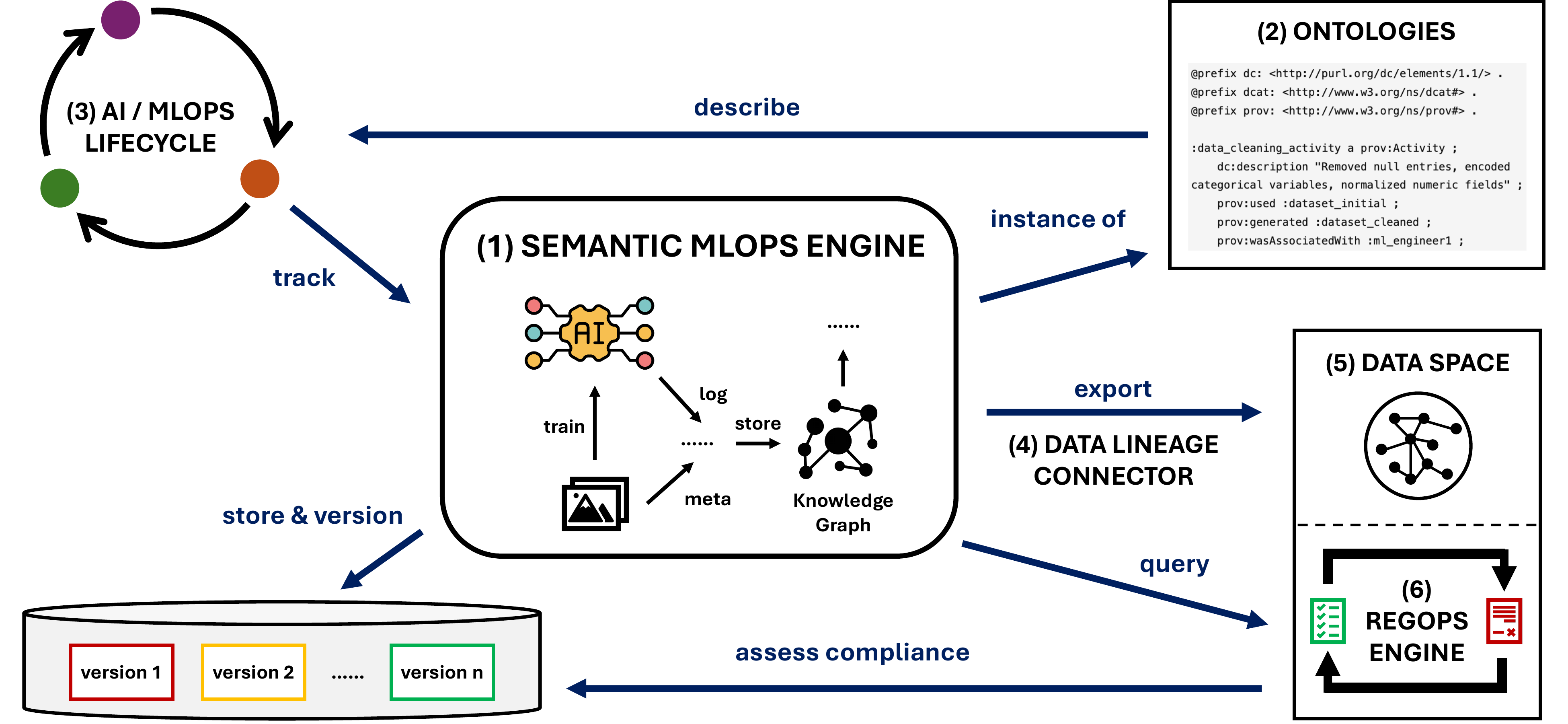}
    \caption{CERTAIN's core components to systematically capture regulatory transparency and enable certification across the AI lifecycle. The architecture highlights the interplay between semantic compliance and infrastructure elements, integrating lifecycle metadata, ontological reasoning, provenance tracking, and compliance validation to ensure auditability and regulatory alignment.}
    \label{fig:certain_framework}
\end{figure}

% The conceptual architecture shown in figure \ref{fig:certain_framework} visually maps CERTAIN’s core components and their relationships. The Semantic MLOps Engine (1) orchestrates the AI/MLOps lifecycle (3) by logging, storing, and versioning artifacts through an integrated version control system. It uses structured knowledge from Ontologies (2) to semantically describe processes and ensure regulatory alignment. The Data Lineage Connector (4) tracks provenance across the lifecycle and interfaces with the Data Space (5), which enables compliant data sharing. The RegOps Engine (6) then queries this ecosystem to assess and validate compliance with applicable regulations.

% To support model certification, a wide range of information must be systematically collected, processed, and stored throughout the entire machine learning (ML) lifecycle. This information varies across the different phases of AI model development and primarily focuses on dataset statistical analysis and versioning, feature engineering and cleaning steps, model architecture, training configurations, evaluation metrics, and deployment details. Additionally, data governance and lineage along with explainability artifacts required for regulatory compliance are also captured. By organizing this information in a structured and auditable manner, the tool enables traceability, reproducibility, and continuous verification against certification standards, facilitating bias and drift detection.

The conceptual architecture shown in Figure~\ref{fig:certain_framework} illustrates CERTAIN’s core components and their interactions. The Semantic MLOps Engine (1) orchestrates the AI/MLOps lifecycle (3) by logging, storing, and versioning artifacts via an integrated version control system, while leveraging structured knowledge from Ontologies (2) to ensure semantic consistency and regulatory alignment. To support model certification, it systematically collects information across the AI lifecycle as well as data governance, lineage, and explainability artifacts required for regulatory compliance, reproducibility, and ongoing verification. A Data Lineage Connector (4) tracks provenance throughout the lifecycle and connects with the Data Space (5) to support secure, compliant data sharing. Finally, the RegOps Engine (6) queries this ecosystem to assess whether certification criteria are met, based on the collected artifacts and lifecycle metadata.

\subsection{Semantic MLOps and Ontology Development\label{sec:semanticframework}}

Recognizing challenges in maintaining consistency, scalability, and standardization in MLOps practices as highlighted in recent studies \cite{zarour2025}, a Semantic MLOps Engine is designed to systematically capture detailed metadata across all stages of the ML lifecycle. This includes data preprocessing, feature engineering, model training, evaluation, deployment, as well as data governance, lineage, and explainability artifacts required for regulatory compliance. By organizing this information in a structured and auditable way, the engine enables traceability, reproducibility, and continuous verification against certification standards. In parallel, metadata on resource utilization such as energy consumption, memory, and compute usage, are collected to support monitoring of environmental impacts. This facilitates the reuse of modular components and aligns with the principles of digital sustainability by embedding environmental accountability and transparency into ML workflows.

Functioning as the “brain” of the system, the Semantic MLOps Engine incorporates legal, ethical, and societal guidelines and links tools for data holders and dataspaces with those for AI providers, buyers, and users. It orchestrates end-to-end workflows and integrates with storage, training, deployment, and governance components to form a certifiable ecosystem. Artifacts are continuously tracked and transformed to be queryable by the RegOps Engine, which validates them against certification criteria and enables the generation of certification reports.

%T5.3 Ontologies to capture and describe the AI lifecycle stages
To systematically capture, trace, and semantically describe the stages of the AI lifecycle, the CERTAIN project adopts a principled ontology engineering methodology \cite{upm76473} grounded in reusability, formal rigor, and regulatory alignment. The ontology development process is driven by competency questions derived from legal obligations (i.e., the EU AI Act), technical standards, and real-world AI engineering practices from the project's pilot applications. It builds upon and extends established vocabularies such as PROV-O, ML-Schema, and emerging responsible AI/accountability ontologies (e.g., the RAInS ontology \cite{naja22}), enabling interoperable descriptions of lifecycle elements such as data sourcing, model development, evaluation metrics, and deployment modalities.

A modular design ensures each AI lifecycle phase is formally represented and traceably linked to compliance-relevant artifacts. 
The developed ontology will be further used by the Semantic MLOps Enginge to structure the collected lifecycle metadata, enabling automated conformity checks, and supporting the generation of machine-actionable documentation for auditability and certification workflows.

\subsection{Infrastructure and Compliance Mechanisms\label{sec:infracompliance}}
Data spaces are emerging as critical infrastructures within the EU Data Strategy, potentially serving as the technical and legal backbone for a new fundamental freedom of data \cite{penedo2024}. Within CERTAIN, data spaces function as secure, interoperable ecosystems that embed regulatory compliance into their architecture. A Data Lineage Connector links the Semantic MLOps infrastructure to data spaces by leveraging semantic web technologies and Knowledge Graphs. This enables detailed provenance tracking and traceability of data throughout the AI lifecycle supporting auditability, accountability, and transparency for all stakeholders.

%T4.4: Design of tools to assess regulatory compliance
%\begin{itemize}
%    \item Short overview of how the compliance assesment works (or could work)
%    \item Concrete outputs and benefits of the chosen methods
%\end{itemize}
%Compliance assessment models designed under CERTAIN automate the validation of AI systems against key regulatory requirements (e.g., fairness, robustness, transparency). These tools operate by systematically generating realistic synthetic test data, revealing potential compliance risks or failures in AI models. The concrete outputs include detailed compliance reports, automated alerts for regulatiory deviations, and actionable recommendations for remedation. Benefits include significantly reduced manual effort for compliance checks, increased accuracy and concistency assessment, and accelerated readiness for certification, ultimately fostering a trustworthy AI environment aligned with EU regulations.
To ensure AI systems meet regulatory requirements, CERTAIN develops compliance assessment models that automate validation against key criteria such as fairness, robustness, and transparency. These tools generate synthetic test data to identify compliance risks, producing detailed reports, alerts, and remediation guidance. This automation reduces manual effort, increases consistency, and accelerates certification readiness contributing to a trustworthy ecosystem aligned with regulations. This includes horizontal EU legislation such as the General Data Protection Regulation (GDPR) and the EU AI Act, as well as vertical, sector-specific regulations (e.g. PSD2 and MiFID II). Starting with the recently enacted EU AI Act, CERTAIN will systematically assess the legal requirements and map them to corresponding technical specifications, enabling the development of tailored software components, such as: (i) Data anonymization tools to support GDPR compliance; (ii) modules for bias detection, mitigation, and Explainable AI to comply with the AI Act, and; (iii) Sector-specific compliance tools aligned with domain-specific regulatory obligations.

%T4.5: Introduction of RegOps worlflows as reusable services within data spaces
%\begin{itemize}
%    \item Short description of reusable RegOps workflow value chains
%\end{itemize}
% CERTAIN aims to design and develop a suite of tools and technologies to enable regulatory-compliant data spaces by design. These solutions will support data holders and consumers for secure and trusted data sharing, in full compliance with applicable regulations. This includes horizontal EU legislation such as the General Data Protection Regulation (GDPR) and the Artificial Intelligence Act (AI Act), as well as vertical, sector-specific regulations (e.g. PSD2 and MiFID II). For each regulation, CERTAIN will systematically assess the legal requirements and map them to corresponding technical specifications, enabling the development of tailored software components, such as: (i) Data anonymization tools to support GDPR compliance; (ii) modules for bias detection, mitigation, and Explainable AI (XAI) to comply with AI Act, and; (iii) Sector-specific compliance tools aligned with domain-specific regulatory obligations.

These components will be integrated into end-to-end regulatory compliance workflows, delivered as configurable and adaptive services within data spaces. An orchestrator engine will enable the coordination of these complex computational workflows and data processing pipelines. All tools and workflows will adhere to the RegOps paradigm and will be demonstrated through a Minimum Viable Data Space implementation. The architecture has been designed for scalability, efficiency, and cost-effectiveness, ensuring streamlined and sustainable compliance with the regulatory landscape.

\section{Current Status, Challenges, and Next Steps}
The CERTAIN project has made tangible progress in laying the groundwork for regulatory-compliant AI development. Initial ontology drafts aligned with EU regulations have been created to capture key lifecycle stages and trustworthiness attributes such as fairness and transparency, with a focus on interoperability and reuse. Furthermore, prototype components of the Semantic MLOps Engine and early RegOps workflows, conceptualized as CI/CD-like compliance pipelines, are under development. These tools are designed from the outset with pilot integration in mind, ensuring applicability across diverse sectors and regulatory contexts.

Key challenges include semantic interoperability across heterogeneous systems and the need to balance formal rigor with practical usability in ontology design. Ensuring alignment between technical artifacts and legal requirements remains a complex task.

Next steps involve integrating these components into seven pilot domains including healthcare, biometrics, energy, finance, and IT to test compliance tools and validate certification procedures. Efforts will also focus on scaling RegOps services across domains, refining orchestration mechanisms, and delivering modular, reusable compliance pipelines to support a certifiable AI ecosystem with EU regulations in mind.

\section{Acknowledgements}
This project has received funding from the European Union's Horizon Europe research and innovation programme [HORIZON-CL4-2024-DATA-01-01] under grant agreement No. 101189650, and  from the Swiss State Secretariat for Education, Research and Innovation (SERI).

\bibliographystyle{plain}
\bibliography{references}

%%%%%%%%%%%%%%%%%%%%%%%%%%%%%%%%%%%%%%%%%%%%%%%%%%%%%%%%%%%%

\end{document}